%% file: main.tex
\documentclass[11pt,a4paper]{article}
\usepackage{naacl2021}

\usepackage{times}
\usepackage{latexsym}
\usepackage{xspace}
\usepackage{enumitem}

\usepackage{url}

\usepackage{booktabs}
\usepackage{diagbox}
\usepackage{multirow}
\usepackage{balance}
\usepackage{amsfonts}
\usepackage{amsmath}
\usepackage{amssymb}
\usepackage{graphicx}
\usepackage{subfigure}
\usepackage{microtype}
\usepackage{xspace}
\usepackage{wrapfig,lipsum}
\usepackage{algorithm}
\usepackage{algorithmicx}
\usepackage{algpseudocode}
\usepackage{makecell}
\usepackage{adjustbox}
\usepackage[T1]{fontenc}
\usepackage{soul}
\usepackage{tabularx}
\usepackage{comment}
\usepackage{todonotes}
\newcommand{\cbb}{\mathbf{c}}
\newcolumntype{Y}{>{\centering\arraybackslash}X}
\newcolumntype{L}{>{\arraybackslash}X}
\newcommand{\cX}{\mathcal{X}}
\newcommand{\cN}{\mathcal{N}}
\algnewcommand\algorithmicinput{\textbf{Input:}}
\algnewcommand\algorithmicoutput{\textbf{Output:}}
\algnewcommand\algorithmicparameter{\textbf{Parameters:}}
\algnewcommand\INPUT{\item[\algorithmicinput]}
\algnewcommand\OUTPUT{\item[\algorithmicoutput]}
\algnewcommand\PARAMETER{\item[\algorithmicparameter]}

\newcommand\DNAME{TAT}

\newcommand{\eat}[1]{\ignorespaces}

\title{
Targeted Adversarial Training for Natural Language Understanding
}

\author{
Lis Pereira\textsuperscript{1}\thanks{~~Equal contribution.}, Xiaodong Liu\textsuperscript{2}$^*$, Hao Cheng\textsuperscript{2}, Hoifung Poon\textsuperscript{2}, Jianfeng Gao\textsuperscript{2}, Ichiro Kobayashi\textsuperscript{1}
 \\ 
  \textsuperscript{1} Ochanomizu University ~~~~~~
  \textsuperscript{2} Microsoft Research \\
  {\tt \{kanashiro.pereira,kobayashi.ichiro\}@ocha.ac.jp} \\
  {\tt \{xiaodl,chehao,hoifung,jfgao\}@microsoft.com}
}

\date{}

\begin{document}
\maketitle
\input{0_abstract}
\input{1_intro}

\input{2_robust_learning}
\input{4_exp}
% \input{5_related_work}
\input{6_discussion}
\input{acknowledgments}

%\appendix
% \input{_appendix}
\bibliography{ref}
\bibliographystyle{acl_natbib}
\clearpage
\appendix
\input{_appendix}
\end{document}

%% file: 0_abstract.tex
% !TEX encoding = UTF-8
% !TEX Root = Main.tex

\begin{abstract}
We present a simple yet effective \textbf{T}argeted \textbf{A}dversarial \textbf{T}raining (\textbf{\DNAME}) algorithm to improve adversarial training for natural language understanding. The key idea is to introspect current mistakes and prioritize adversarial training steps to where the model errs the most. Experiments show that {\DNAME} can significantly improve accuracy over standard adversarial training on GLUE and attain new state-of-the-art zero-shot results on XNLI. Our code will be released at: \url{https://github.com/namisan/mt-dnn}.  
\end{abstract}

%% file: 1_intro.tex
% !TEX encoding = UTF-8
% !TEX Root = Main.tex

\section{Introduction}
\label{sec:intro}
%Large-scale pre-trained language models such as BERT \cite{devlin2018bert} have achieved great success in a wide range of natural language processing (NLP) tasks. These models are often fine-tuned on downstream tasks using a transfer learning approach, where the top layer of the language model is replaced by a task specific sub-network, and then the new model is further trained with the low-resource data of the target task. However, due to the limited data from the target tasks and the extremely high complexity of the pre-trained model, aggressive fine-tuning can easily make the adapted model overfit the data of the target task, making it unable to generalize well on unseen data (Jiang et al., 2019). Moreover, some researchers have shown that such pre-trained models are vulnerable to adversarial attacks \cite{jin2019bertrobust}.

Adversarial training has proven effective in improving model generalization and robustness in computer vision \cite{madry2017pgd,goodfellow2014explaining} and natural language processing (NLP) \cite{zhu2019freelb,jiang2019smart,cheng-etal-2019-adv-nmt,liu2020alum,pereira2020alic,cheng2020posterior}. 
\input{tables/cfmatrix}
It works by augmenting the input with a small perturbation to steer the current model prediction away from the correct label, thus forcing subsequent training to make the model more robust and generalizable. Aside from some prior work in computer vision \cite{dong2018boosting,tramer2017ensemble}, most adversarial training approaches adopt {\em non-targeted} attacks, where the model prediction is not driven towards a specific incorrect label. In NLP, the cutting-edge research in adversarial training tends to focus on making adversarial training less expensive (e.g., by reusing backward steps in FreeLB \cite{zhu2019freelb}) or regularizing rather than replacing the standard training objective (e.g., in virtual adversarial training (VAT) \cite{jiang2019smart}).

By contrast, in this paper, we investigate an orthogonal direction by augmenting adversarial training with introspection capability and adopting {\em targeted} attacks to focus on where the model errs the most. We observe that in many NLP applications, the error patterns are non-uniform. For example, in the MNLI development set (in-domain), standard fine-tuned BERT model tends to misclassify a non-neutral instance as ``neutral'' more often than the opposite label (Figure~\ref{fig:cmt} top). We thus propose {\em Targeted Adversarial Training} (TAT), a simple yet effective algorithm for adversarial training. For each instance, instead of taking adversarial steps {\em away} from the gold label, TAT samples an incorrect label proportional to how often the current model makes the same error in general, and takes adversarial steps {\em towards} the chosen incorrect label. 
To our knowledge, this is the first attempt to apply targeted adversarial training to NLP tasks. 
In our experiments, this leads to significant improvement over standard non-adversarial and adversarial training alike. For example, in the MNLI development set, TAT produced an accuracy gain of 1.7 absolute points (Figure~\ref{fig:cmt} bottom). On the overall GLUE benchmark, TAT outperforms state-of-the-art non-targeted adversarial training methods such as FreeLB and VAT, and enables the BERT\textsubscript{BASE} model to perform comparably to the BERT\textsubscript{LARGE} model with standard training. The benefit of TAT is particularly pronounced in out-domain settings, such as in zero-shot learning in natural language inference, attaining new state-of-the-art cross-lingual results on XNLI.

%% file: tables/cfmatrix.tex
\begin{figure}[htb!]
\centering  
\subfigure[BERT with standard fine-tuning]{
\adjustbox{trim={0.0\width} {0.04\height} {0.\width} {0.02\height},clip}{
\includegraphics[width=0.9\linewidth]{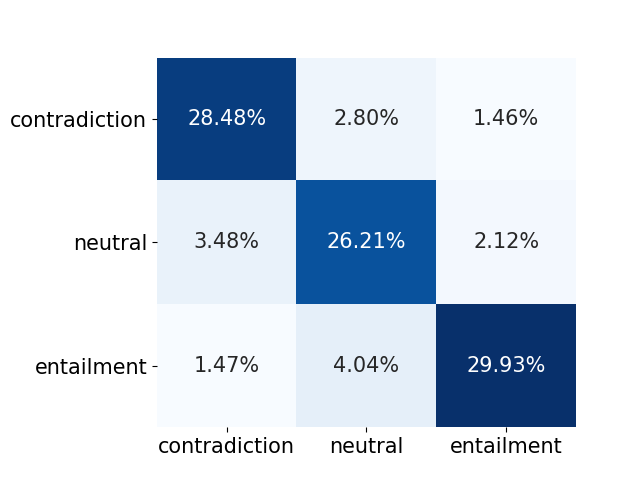}
}}
%\vspace{-4mm}
\subfigure[BERT with TAT fine-tuning]{
\adjustbox{trim={0.0\width} {0.04\height} {0.\width} {0.02\height},clip}{
\includegraphics[width=0.9\linewidth]{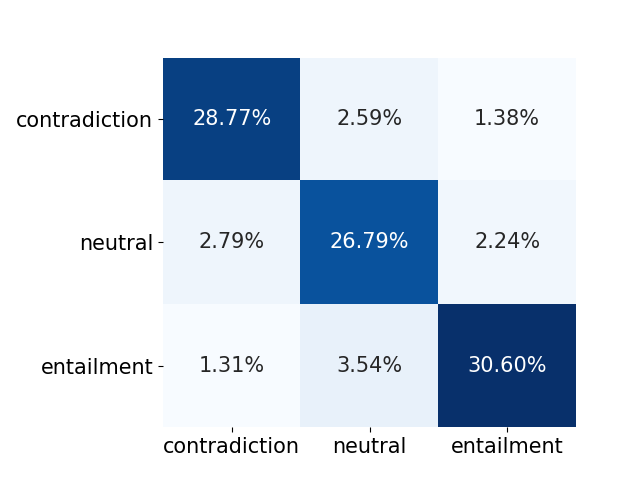}
}}
	\caption{
 Comparison of confusion matrices on MNLI development set (in-domain). X-axis and Y-axis represent the predicted and gold labels, respectively. TAT produces an accuracy gain of 1.7 absolute points.
	}
%\vspace{-4mm}
	\label{fig:cmt}
\end{figure}

%% file: 2_robust_learning.tex
% !TEX encoding = UTF-8
% !TEX Root = Main.tex
\section{Targeted Adversarial Training (TAT)}
\label{sec:robust_learning}
In this paper, we focus on fine-tuning BERT models \cite{devlin2018bert} in our investigation of targeted adversarial training, as this approach has proven very effective for a wide range of NLP tasks. 

The training algorithm seeks to learn a function $f(x; \theta): x \rightarrow C$ as parametrized by $\theta$, where $C$ is the class label set.
Given a training dataset $D$ of input-output pairs $(x,y)$ and the loss function $l(.,.)$ (e.g., cross entropy), the standard training objective would minimize the empirical risk: 
\[ \min_{\theta} \mathbb{E}_{(x, y)\sim D}[l(f(x; \theta), y)]. \]  
%\[ \min_{\theta} \mathbb{E}_{(x, y)\sim D}[l(f(x; \theta), y)]. \]  
%While this is effective to train a classifier, it usually suffers from overfitting and poor generalization to unseen cases.

By contrast, in adversarial training, as pioneered in computer vision \cite{goodfellow2014explaining,hsieh2019robust-self-att,madry2017pgd,jin2019bertrobust}, the input would be augmented with a small perturbation that maximize the adversarial loss:
\[ \min_{\theta} \mathbb{E}_{(x, y)\sim D}[\max_{ \delta} l(f(x + \delta; \theta), y)], \]  
where the inner maximization can be solved by projected gradient descent \cite{madry2017pgd}. 

Recently, adversarial training has been successfully applied to NLP as well \cite{zhu2019freelb,jiang2019smart,pereira2020alic}. In particular, 
FreeLB \cite{zhu2019freelb} leverages the {\em free adversarial training} idea \cite{shafahi2019freeat} by reusing the backward pass in gradient computation to carry out inner ascent and outer descent steps simultaneously. SMART \cite{jiang2019smart} instead regularizes the standard training objective using {\em virtual adversarial training} \cite{miyato2018virtual}:
\begin{equation}
%\vspace{-2mm}
\begin{aligned}
  \min_{\theta} \mathbb{E}_{(x, y)\sim D}[l(f(x; \theta), y) + \\ 
  \alpha  \max_{
\delta} l(f(x+\delta; \theta), f(x; \theta))]
\end{aligned}
%\vspace{-1mm}
\label{eq:alum}
\end{equation}
Effectively, the adversarial term encourages smoothness in the input neighborhood, and $\alpha$ is a hyperparameter that controls the trade-off between standard errors and adversarial errors. %In our experiments, we simply set $\alpha$ to 1.

\input{tables/alg}

\input{tables/adv_comp}

In standard adversarial training, the algorithm simply tries to perturb the input $x$ away from the gold label $y$ given the current parameters $\theta$. It is agnostic to which incorrect label $f(x)$ might be steered towards. By contrast, in Targeted Adversarial Training (TAT), we would explicitly pick a target $y_t\ne y$ and try to steer the model towards $y_t$. Intuitively, we would like to focus training on where the model currently errs the most. We accomplish this by keeping a running tally of $e(y,y_t)$, which is the current expected error of predicting $y_t$ when the gold label is $y$, and sample $y_t$ from $C_{\setminus y}=C-\{y\}$ in proportion to $e(y,y_t)$. See Algorithm~\ref{algo:main} for details. 
TAT can be applied to the original adversarial training or virtual adversarial training alike. In this paper, we focus on adapting virtual adversarial training (VAT) \cite{jiang2019smart}.
The two lines in blue color are the only change from VAT. 
We initialize $e(y,y_t)$ with uniform distribution and update them in each epoch. We conducted an oracle experiment where $e(y,y_t)$ was taken from the confusion matrix from standard training and found that it performed similarly as our online version. 

It is more challenging to apply TAT to regression tasks, as we would need to keep track of a continuous error distribution. To address this problem, we quantize the value range into ten bins and apply TAT similarly as in the classification setting (once a bin is chosen, a value is sampled uniformly within).

%%%%%%%%%%%%%%%%%%%%%%%%%%%%%%%%%%%%%%%%%%%%%%%%%
\eat{
With the goal of developing more robust models, adversarial training has primarily been used in computer vision \cite{goodfellow2014explaining,hsieh2019robust-self-att,madry2017pgd,jin2019bertrobust}. The key idea is to modify the training objective by applying small perturbations to input images that maximize the adversarial loss:
$\min_{\theta} \mathbb{E}_{(x, y)\sim D}[\max_{ \delta} l(f(x + \delta; \theta), y)]$, 
where the inner maximization can be solved by running a number of projected gradient descent steps \cite{madry2017pgd}. 
%However, models trained with adversarial learning have been found to have a degradation in generalization \cite{tsipras2018robustness,zhang2019theoretically}. 
In NLP, in order to improve both the model generalization and robustness, FreeLB \cite{zhu2019freelb} and SMART \cite{jiang2019smart} utilize adversarial training algorithms that apply perturbations at the embedding level \cite{zhu2019freelb,jiang2019smart,pereira2020alic}.

\input{tables/alg}
\input{tables/adv_comp}
Although these adversarial training algorithms greatly improve model performance and generalization, such attacks are aimless. In many real-world applications, there are always existing classes or categories on which the learned classifier has a higher error rate. For instance, given a pair of sentence with a correct label ``contradition'' on natural language inference, it is easier to misclassify to ``neutral'' instead of ``entailment''. Inspired by targeted attack learning \cite{dong2018boosting,tramer2017ensemble}, we first incorporate such prior knowledge into adversarial training in order to reduce error in these classes. The detailed algorithm is shown in Algorithm~\ref{algo:main}. Comparing with standard adversarial training, TAT first samples a label from the class label set (excluding the correct label) \footnote{{\DNAME} works well on multi-class classification where a confusion matrix encodes much richer global information, and in case of binary classification, it always attacks the incorrect label which is similar to attach the correct label.} where we can flexibly use different prior knowledge, e.g., \textcolor{blue}{the confusion matrix \footnote{\textcolor{blue}{We can also sample a small part of training data to estimate the confusion matrix online where the initial distribution is set to uniform, and it is updated after each epoch. In practice, we find that both approaches give the similar performance on the representative MNLI task.}} from the old model using the standard training, where its diagonal is set to 0, the other values renormalized for a random \textit{choice} of the targeted label}, as shown in line 4. Then the adversarial sample is estimated by the opposite direction as in line 7. For regression tasks, we create several bins to mimic the classification task to sample a different bin other than the correct one. At last, following \cite{jiang2019smart,miyato2018virtual}, the adversarial regularizer is added to standard training objective (e.g., cross-entropy between the correct label and prediction). 
}

%% file: tables/alg.tex
\begin{algorithm}[!tb]
\caption{{\DNAME}}\label{algo:main}
\begin{algorithmic}[1]
	\INPUT $T$: the total number of iterations, $\cX=\{(x_1, y_1),...,(x_n, y_n)\}$:  the dataset,
	$f(x; \theta)$: the machine learning model parametrized by $\theta$, $\sigma^2$: the variance of the random initialization of perturbation $\delta$, $\epsilon$: perturbation bound, $K$: the number of iterations for perturbation estimation, $\eta$: the step size for updating perturbation, $\tau$: the global learning rate, $\alpha$: the smoothing proportion of adversarial training in the augmented learning objective, $\Pi$: the projection operation and $C$: the classes.  

	\For{$t=1,..,T$}
	\For{$(x,y) \in$ $\mathcal{X}$}
	    \State $\delta \sim \cN(0,\sigma^2I)$
        \State \textcolor{blue}{$y_t = sample(C_{\setminus y})$} 
    	\For{$m=1,..,K$}
    	    \State $g_{adv} \leftarrow \nabla_{\delta} l(f(x + \delta; \theta), y_t)$
    	    \State $\delta \leftarrow \Pi_{\|\delta\|_{\infty} \leq \epsilon} (\textcolor{blue}{\delta - \eta g_{adv}})$  
    	\EndFor
	    \State $g_{\theta} \leftarrow \nabla_{\theta} l(f(x; \theta), y)$  
	    
	    $\qquad\quad+ \alpha \nabla_{\theta}l(f(x; \theta), f(x + \delta; \theta))$ 
	    \State $\theta \leftarrow \theta - \tau g_{\theta}$
	\EndFor
	\EndFor
		\OUTPUT $\theta$
\end{algorithmic}
\end{algorithm}

%% file: tables/adv_comp.tex
\begin{table*}[!htb]
%\vspace{-0.1in}
	\begin{center}
		\begin{tabular}{@{\hskip2pt}l@{\hskip2pt}|@{\hskip2pt}c@{\hskip2pt}|@{\hskip1pt}c@{\hskip2pt}|@{\hskip2pt}c@{\hskip2pt}|@{\hskip2pt}c@{\hskip2pt}|@{\hskip2pt}c@{\hskip2pt}|@{\hskip2pt}c@{\hskip2pt}|@{\hskip2pt}c @{\hskip1pt}|@{\hskip1pt}c@{\hskip2pt}|@{\hskip2pt}c@{\hskip2pt}}
		\toprule
		\bf Methods            &MNLI-{m/mm}        &QQP                &RTE            &QNLI           &MRPC               &CoLA           &SST          &STS-B & Average \\ 
			  %\textbf{BERT\textsubscript{BASE}} (uncased)
			  &Acc                &Acc/F1             &Acc            &Acc            &Acc/F1             &Mcc            &Acc            &P/S Corr & Score      \\ \midrule
         Standard (BERT\textsubscript{LARGE})\textsuperscript{dev}   &86.3/86.2 & 91.3/88.4 & 71.1 & 92.4 & 85.8/89.5 &61.8 &93.5 & 89.6/89.3 &84.0\\ %\hline
	         Standard (BERT\textsubscript{LARGE})\textsuperscript{test}   &86.7/85.9 & 72.1/89.3 & 70.1 & 92.7 & 85.4/89.3 &60.5 &94.9 & 87.6/86.5 & 82.4\\ \hline

		%	STD   &84.4/-             &-                  &-              &88.4           &-             &-              &92.7           &-\\ \hline
			Standard\textsuperscript{dev}            &84.5/84.4          &90.9/88.3          &63.5           &91.1           &84.1/89.0          &54.7           &92.9           &89.2/88.8 &81.5\\  %\hline
            {FreeLB}\textsuperscript{dev}                     &85.4/85.5 &91.4/88.4 &70.4 & 91.5 &86.2/90.3 &\textbf{59.1} &93.2  &89.7/89.1 &83.5\\ %\hline
            VAT\textsuperscript{dev}                     &85.5/85.7 &91.5/88.5 &71.2 & 91.7 &87.7/91.3 &58.2 &93.3  &90.0/89.4&83.7 \\ %\hline
        %    {TextAT}                     &85.7/85.8 &91.6/88.9 &\textbf{75.2} & 92.4 &88.0/91.6 &62.0 &93.7  &90.0/89.6 \\ \hline             
            
            {\DNAME}\textsuperscript{dev}                   &\textbf{86.2/85.9} & \textbf{91.8/89.1}&\textbf{72.6}&\textbf{92.2}&\textbf{88.2/91.5}& 58.5 & \textbf{93.6}& \textbf{90.8/89.6} &\textbf{84.2}\\ \hline
		 Standard\textsuperscript{test}               &84.6/83.4          &71.2/89.2          &66.4           &90.5           &84.8/88.9         &52.1           &93.5           &87.1/85.8 &80.0\\  %\hline

        {\DNAME}\textsuperscript{test}                   &\textbf{85.8/84.8} &\textbf{72.8/89.6}&\textbf{69.7}&\textbf{92.4}&\textbf{88.2/91.1}& \textbf{59.8} & \textbf{94.5}& \textbf{89.7/89.0} &\textbf{82.8}\\
        \bottomrule
\end{tabular}
	\end{center}
	%\vspace{-0.15in}
	\caption{Comparison of standard and adversarial training methods on GLUE. All rows except the top two use standard BERT\textsubscript{BASE} model. The GLUE test results are scored using the GLUE evaluation server. Note that the test results of Standard including BERT\textsubscript{BASE} and BERT\textsubscript{LARGE} are taken from https://gluebenchmark.com/leaderboard.
	}
	%\vspace{-5mm}
	\label{tab:glue_dev}
\end{table*}

%% file: 4_exp.tex
\section{Experiments}
\label{sec:exp}
We compare targeted adversarial training (TAT) with standard training and state-of-the-art adversarial training methods such as FreeLB \cite{zhu2019freelb} and VAT \cite{miyato2018virtual,jiang2019smart}. 
We use the standard uncased BERT\textsubscript{BASE} model \cite{devlin2018bert}, unless noted otherwise. 
Due to the additional overhead incurred during training, adversarial methods are somewhat slower than standard training. Like VAT, TAT requires an additional $K$ adversarial steps compared to standard training. In practice, $K=1$ suffices for TAT and VAT, so they are just slightly slower (roughly 2 times compared to standard training). FreeLB, by contrast, typically requires 2-5 steps to attain good performance, so is significantly slower. % (2-5 times compared to standard training depends on the number of steps).

\subsection{Implementation Details}
\label{sec:param}
Our implementation is based on the MT-DNN toolkit \cite{mtdnn2020demo}. We follow the default hyperparameters used for fine-tuning the uncased BERT base model \cite{devlin2018bert, mtdnn2020demo}. 
Specifically, we use $0.1$ for the dropout rate except 0.2 for MNLI, $0.01$ for the weight decay rate and the Adamax \cite{kingma2014adam} optimizer with the default Lookahead \cite{zhang2019lookahead} to stabilize training. 
We select the learning rate from $\{5\mathrm{e}{-5}, 1\mathrm{e}{-4}\}$ for all the models. The maximum training epoch is set to 6, and the we follow \cite{jiang2019smart} to set adversarial training hyperparameters: $\epsilon=1\mathrm{e}{-5}$ and $\eta=1\mathrm{e}{-4}$. In our experiments, we simply set $\alpha=1$ in Eq 1.

\subsection{Standard GLUE Evaluation}
\label{subsec:glue}
%\vspace{-2mm}
We first compare adversarial training methods on the standard GLUE benchmark \cite{wang2018glue}. See \autoref{tab:glue_dev} for the results \footnote{Due to restriction on the number of submissions by the GLUE organizers, we only compared {\DNAME} with the published results from \cite{devlin2018bert} on the test set.}.
{\DNAME} consistently outperforms both standard training and the state-of-the-art adversarial training methods of FreeLB and VAT. 
%Noting that on MNLI {\DNAME} obtains approximately 0.5\% improvement on average of both in/out-domain over VAT. 
Remarkably, BERT\textsubscript{BASE} with targeted adversarial training performs on par with BERT\textsubscript{LARGE} with standard training overall, and outperforms the latter by a large margin on tasks with smaller datasets such as RTE, MRPC and STS-B, which illustrates the benefit of TAT in improving model generalizability.

\input{tables/zero_shot}
\input{tables/xnli}

\subsection{Zero-Shot Learning on Natural Language Inference}
\label{subsec:nli}

Next, we compare standard and adversarial training in generalizability to out-domain datasets. 
Specifically, we fine-tune BERT\textsubscript{BASE} on the MNLI training data and evaluate it on various natural language inference test sets: HANS \cite{mccoy2019hans}, SNLI \cite{snli2015}, SciTail \cite{scitail}, MeNLI \cite{romanov2018mednli}. 
See \autoref{tab:domain_shift} for the results. TAT substantially outperforms standard training and state-of-the-art adversarial training methods. Interestingly, the gains are particularly pronounced on the two hardest datasets, HANS and MedNLI. 
HANS used heuristic rules to identify easy instances for MNLI-trained BERT models and introduced modifications to make them harder. 
MedNLI is from the biomedical domain, which is substantially different from the general domain of MNLI. This provides additional evidence that targeted adversarial training is especially effective in enhancing generalizability in out domains.

\subsection{Zero-Shot Learning on Cross-Lingual Natural Language Inference}
\label{subsec:nli}

We also conducted zero-shot evaluation in the cross-lingual setting by comparing standard and adversarial training on XNLI \cite{conneau2018xnli}. 
Specifically, a cross-lingual language model is fine-tuned using the English NLI dataset and then tested on datasets of other languages. Following \newcite{conneau2019xlmr}, we used the pre-trained XLM-R large model in our experiments, and compare targeted adversarial training (XLM-R+TAT) with  state-of-the-art systems that use standard training (XLM-R) and adversarial training (XLM-R+R3F/R4F) \cite{aghajanyan2020rf}, as well as another state-of-the-art language model InfoXLM \cite{chi2020infoxlm}. 
To ensure fair comparison, we also report the results from our reimplementation of XLM-R \cite{conneau2018xnli} (XLM-R\textsubscript{Reprod}).
See \autoref{tab:xnli} for the results. 
Targeted adversarial training (TAT) demonstrates a clear advantage in improving zero-shot transfer learning across languages, especially for languages most different from English, such as Urdu. Overall, TAT produces a new state-of-the-art result of 81.7\% over 15 languages on XNLI.
%%%%%%%%Figures%%%%%%%%%%%%

\subsection{Analysis}

\input{tables/agreement}

As we have seen in Figure~\ref{fig:cmt} earlier, {\DNAME} reduces the errors across the board on MNLI development set. 
To understand how {\DNAME} improves performance, we conducted a more detailed analysis by subdividing the dataset based on the degree of human agreement. Here, there are three label classes and each sample instance has 5 human annotations. The samples can be divided into four categories: 
5-0-0, 4-1-0, 3-2-0, 3-1-1. E.g., 3-1-1 signifies that there are three votes for one label and one for each of the other two labels. In Figure~\ref{fig:agree}, we see that {\DNAME} outperforms the baseline consistently over all categories,  with higher improvement on the more ambiguous samples, especially for out-domain samples. This suggests that {\DNAME} is most helpful for the challenging instances that exhibit higher ambiguity and are more different from training examples. 

%We also visualize the training loss landscape for standard training and {\DNAME}. Standard training renders a much more rugged loss surface, where {\DNAME} produces smoother surface with wider and flatter basins as shown in Figure~3.

% \section{Loss Landscape}
%  \label{sec:lossland}
%  We also visualize the loss landscape of both the standard training and {\DNAME}, shown in Figure~\ref{fig:loss-surface}. {\DNAME} has a wider and flatter loss surface indicating a better generalization \cite{hochreiter1997flat,hao2019vbert,li2018landscape}. 
 
\input{tables/losssurface}
We also visualize the loss landscape of both the standard training and {\DNAME}, shown in Figure~\ref{fig:loss-surface}. {\DNAME} has a wider and flatter loss surface, which generally indicates better generalization \cite{hochreiter1997flat,hao2019vbert,li2018landscape}. 
 
% \input{tables/losssurface}
%\input{tables/robustness}

%% file: tables/zero_shot.tex
\begin{table}[t]
    \centering
    \begin{tabular}{l|c|c|c|c}
    \hline
\toprule    
         \multirow{2}{*}{\bf Method} 
          & HANS & SNLI & SciTail & MedNLI  \\
          & Acc & Acc &Acc & Acc\\ \hline
          Standard &55.4	&80.1 &77.3 & 43.2   \\ 
          FreeLB &62.0	&80.5 &78.6  &56.8 \\ 
          VAT &62.5	&80.8 &78.5   &58.1\\ 
          TAT & \bf 65.8 & \bf 81.0 & \bf 78.8 & \bf 60.6\\
\bottomrule
    \end{tabular}
	%\vspace{-2mm}
    \caption{Comparison of standard and adversarial training in zero-shot evaluation on various natural language inference datasets, where the standard BERT\textsubscript{BASE} model is fine-tuned on the MNLI training data.}
    	%\vspace{-4mm}
    \label{tab:domain_shift}
\end{table}

%% file: tables/xnli.tex
%{'all': 81.67265469061876, 'ar': 80.67864271457086, 'bg': 84.9500998003992, 'de': 83.87225548902195, 'el': 83.67265469061877, 'en': 89.30139720558881, 'es': 85.72854291417165, 'fr': 84.17165668662675, 'hi': 78.42315369261476, 'ru': 82.09580838323353, 'sw': 74.05189620758483, 'th': 79.74051896207584, 'tr': 80.95808383233532, 'ur': 75.12974051896208, 'vi': 81.27744510978044, 'zh': 81.0379241516966}

\begin{table*}[!htb]
	%\vspace{-2mm}
	\begin{center}
		\begin{tabular}{@{\hskip2pt}l @{\hskip2pt} |@{\hskip2pt}c @{\hskip2pt} |@{\hskip2pt}  c @{\hskip2pt} |@{\hskip2pt}  c @{\hskip2pt} |@{\hskip2pt} c @{\hskip2pt} |@{\hskip2pt}  c @{\hskip2pt} |@{\hskip2pt}  c@{\hskip2pt} |@{\hskip2pt}  c @{\hskip2pt} |@{\hskip2pt} c@{\hskip2pt} |@{\hskip2pt}  c@{\hskip2pt} | @{\hskip2pt} c@{\hskip2pt} |@{\hskip2pt}  c@{\hskip2pt} |  c@{\hskip2pt} |  c@{\hskip2pt} |  c@{\hskip2pt} |  c@{\hskip2pt} |  @{\hskip2pt}c@{\hskip2pt} }
		% \begin{tabular}{@{\hskip2pt}l@{\hskip2pt}|c|c|c|c|c|c|c|c|c|c|c|c|c|c|c|c }
		\toprule
		\bf Model     &en    &fr    &es    &de    &el    &bg    &ru    &tr    &ar    &vi    &th    &zh    &hi    &sw    &ur    &Avg.\\
		\toprule
XLM-R                       &89.1    &84.1   &85.1   &83.9   &82.9   &84.0   &81.2   &79.6   &79.8   &80.8   &78.1   &80.2  &76.9   &73.9   &73.8  &80.9 \\
XLM-R\textsubscript{Reprod} &88.1    &83.6   &84.1   &83.0   &82.6   &83.8   &81.7   &80.7   &80.4   &80.7   &78.9   &80.1  &77.8   &74.2   &74.0  &80.9  \\ \hline
XLM-R+R3F                   &89.4    &84.2   &85.1   &83.7   &83.6   &84.6   &82.3   &80.7   &80.6   &81.1   &79.4   &80.1  &77.3   &72.6   &74.2  &81.2 \\
XLM-R+R4F                   &89.6    &\textbf{84.7}    &85.2    &\textbf{84.2}    &83.6    &84.6    &\textbf{82.5}    &80.3    &80.5    &80.9    &79.2    &80.6    &78.2    &72.7    &73.9    &81.4 \\
InfoXLM    &\textbf{89.7}    &84.5    &85.5    &84.1    &83.4    &84.2    &81.3    &80.9    &80.4    &80.8    &78.9    &80.9    &77.9    &\textbf{74.8}    &73.7    &81.4 \\ \hline
\textbf{XLM-R+{\DNAME}}    &89.3      &84.2    &\textbf{85.7}    &83.9    &\textbf{83.7}    &\textbf{85.0}    &82.1    & \textbf{81.0}   &\textbf{80.7}    &\textbf{81.3}    &\textbf{79.7}    &\textbf{81.0}    &\textbf{78.4}    &74.1    &\textbf{75.1}  &\textbf{81.7} \\
        \bottomrule
\end{tabular}
	\end{center}
		%\vspace{-4mm}
%	\caption{Comparison of TAT and prior state of the art in zero-shot cross-lingual learning on the XNLI test set.}
	\caption{Comparison of targeted adversarial training (TAT) and prior state of the art in zero-shot cross-lingual learning on the XNLI test set.}
	%\vspace{-4mm}
	\label{tab:xnli}
\end{table*}

%% file: tables/agreement.tex
\begin{figure}[htb!]
\centering  
\subfigure[MNLI Development (in-domain)]{
\adjustbox{trim={0.0\width} {0.07\height} {0.\width} {0.\height},clip}{
\includegraphics[width=0.95\linewidth]{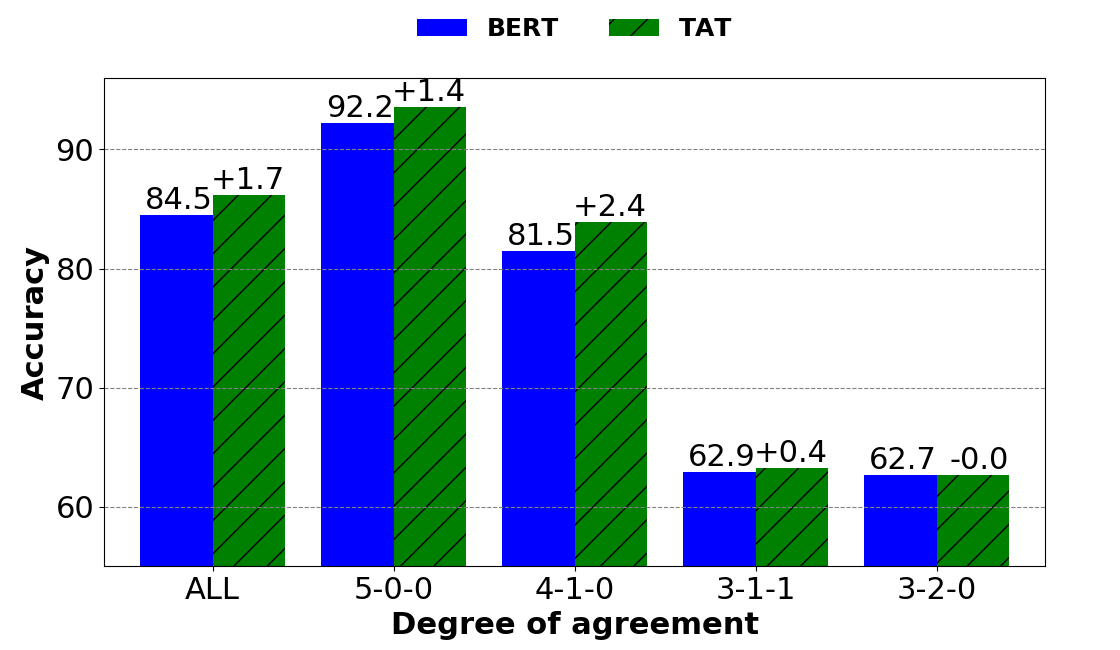}}
}
%\vspace{-5mm}
\subfigure[MNLI Development (out-domain)]{
\adjustbox{trim={0.0\width} {0.07\height} {0.\width} {0.1\height},clip}{
\includegraphics[width=0.99\linewidth]{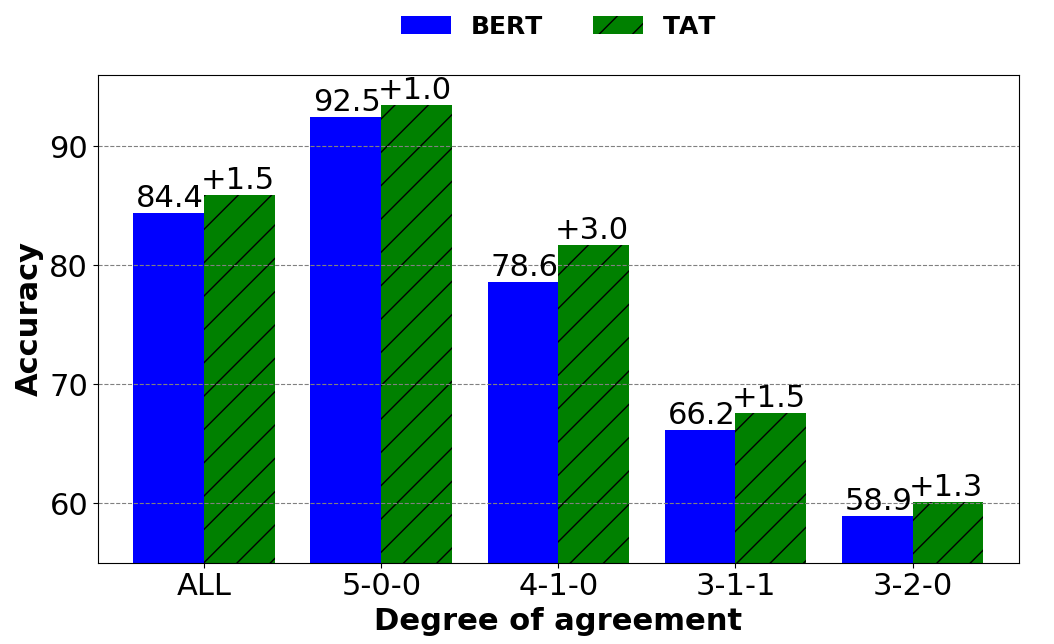}}
}
	\caption{
 Comparison of standard and targeted adversarial training on MNLI, subdivided per agreement.
	}
%\vspace{-5mm}
	\label{fig:agree}
\end{figure}

%% file: tables/losssurface.tex
\begin{figure}[htb!]
\centering  
\subfigure[Loss surface of traditional training]{
\adjustbox{trim={0.0\width} {0.01\height} {0.\width} {0.02\height},clip}{
\includegraphics[width=0.9\linewidth]{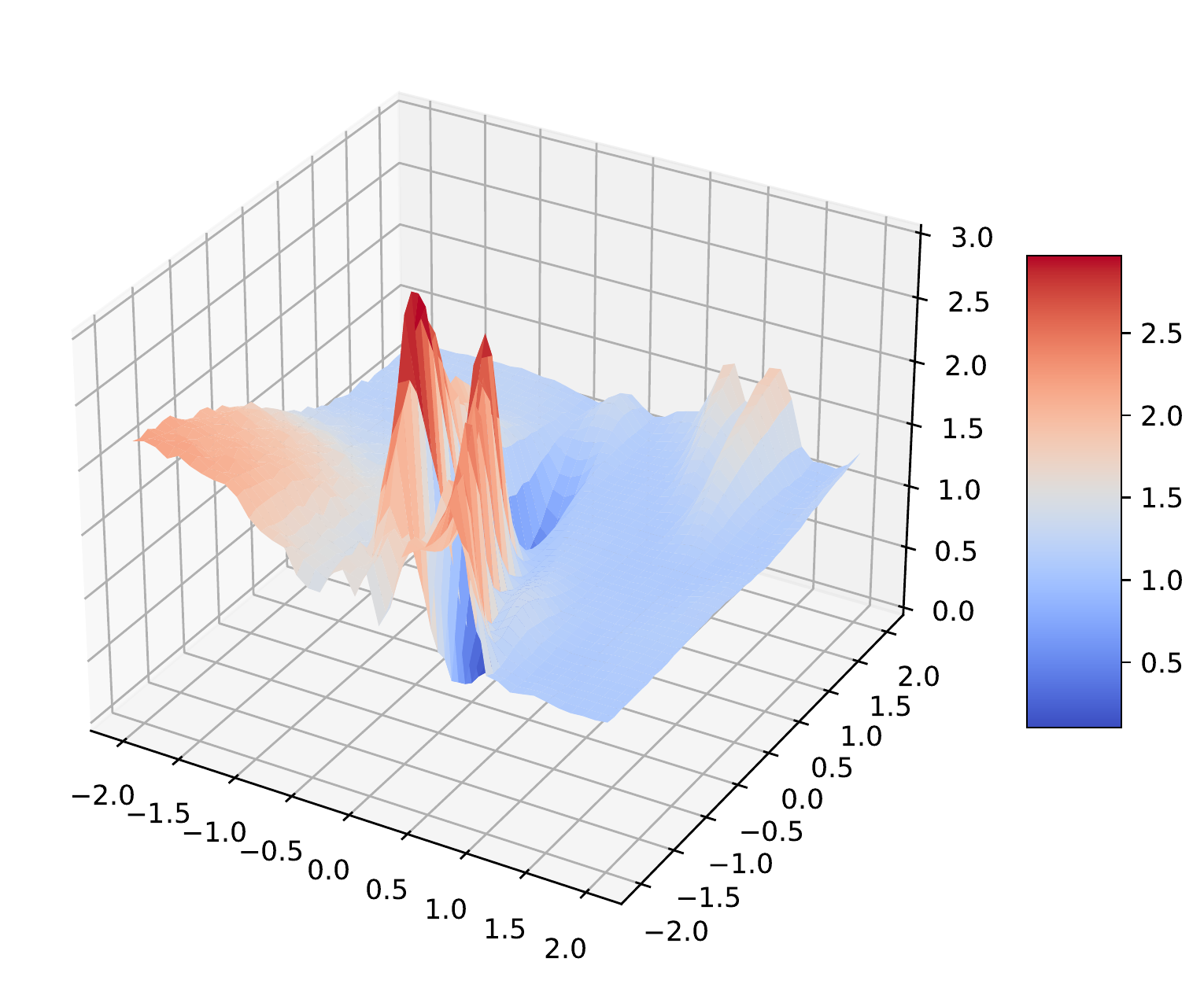}
}}
%\vspace{-4mm}
\subfigure[Loss surface of TAT]{\includegraphics[width=0.9\linewidth]{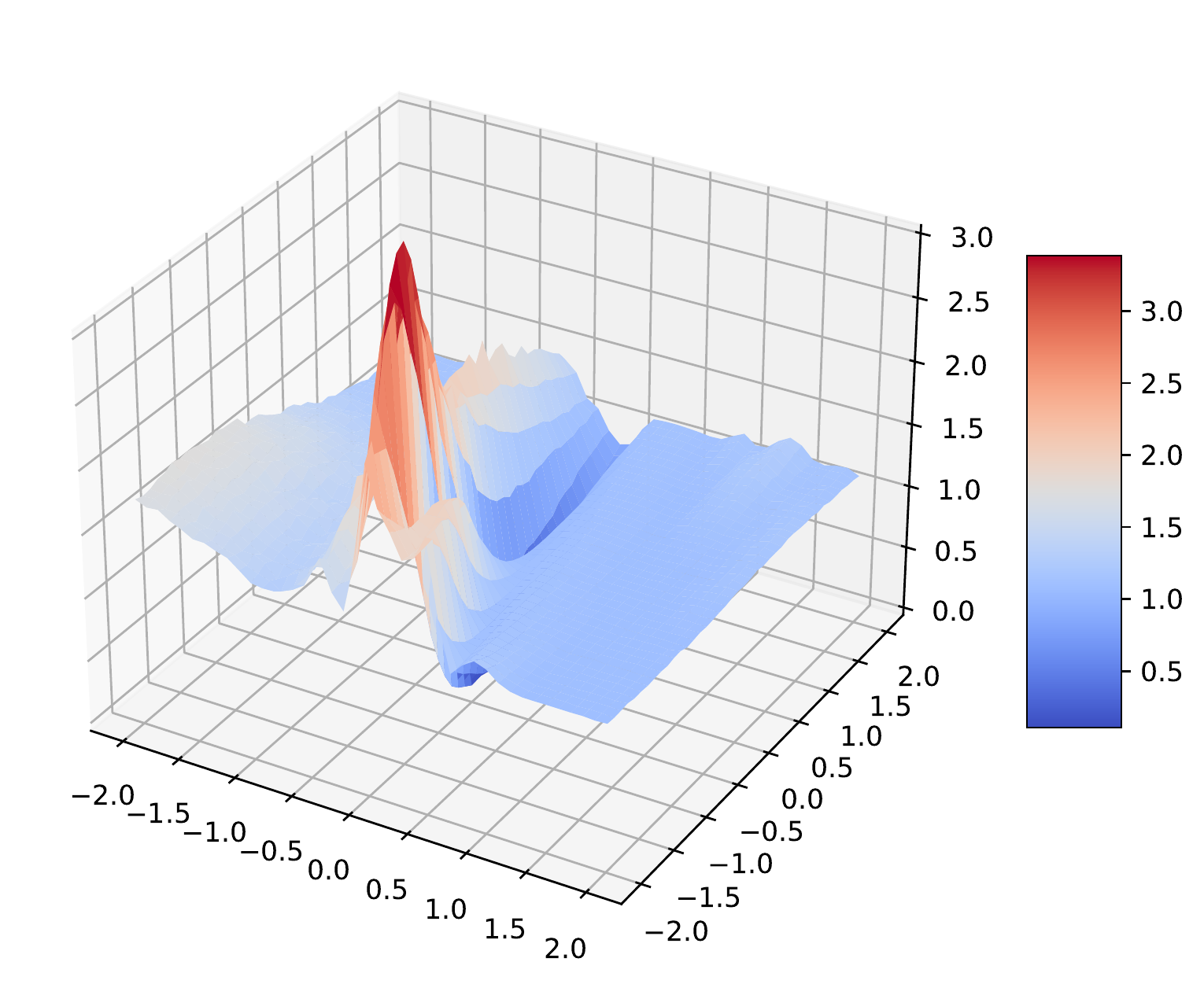}}
	\caption{
Training loss surfaces of traditional training vs TAT on MNLI.
	}
%	\vspace{-4mm}
	\label{fig:loss-surface}
\end{figure}

%% file: 6_discussion.tex
% !TEX encoding = UTF-8
% !TEX Root = Main.tex
%\vspace{-2mm}
\section{Conclusion}
\label{sec:conclusion}
%\vspace{-2mm}
We present the first study to apply targeted attacks in adversarial training for natural language understanding. Our TAT algorithm is simple yet effective in improving model generalizability for various NLP tasks, especially in zero-shot learning and for out-domain data. 
Future directions include: applying TAT in pretraining and other NLP tasks e.g., sequence labeling, exploring alternative approaches for target sampling.

%% file: acknowledgments.tex
\section*{Acknowledgments}
We thank Microsoft Research Technology Engineering team for setting up GPU machines. We also thank the anonymous reviewers for valuable discussions.

%% file: _appendix.tex
\section{NLU Benchmarks}
\label{sec:appendix}
\input{tables/dataset}
The NLU benchmarks used in our experiments, i.e. GLUE benchmark \cite{wang2018glue}, SNLI \cite{snli2015}, SciTail \cite{scitail}, HANS \cite{mccoy2019hans}, MedNLI \cite{romanov2018mednli} and XNLI \cite{conneau2018xnli}, 
are briefly introduced in the following sections. Table~\ref{tab:datasets} summarizes the information of these tasks. In the experiments, GLUE is used for the normal setting, while the other datasets are used for the zero-shot setting.

\noindent $\bullet$ \textbf{GLUE}. The General Language Understanding Evaluation (GLUE) benchmark is a collection of nine natural language understanding (NLU) tasks. As shown in Table~\ref{tab:datasets},
it includes question answering~\cite{squad1}, linguistic acceptability~\cite{cola2018}, sentiment analysis~\cite{sst2013}, text similarity~\cite{sts-b2017}, paraphrase detection~\cite{mrpc2005}, and natural language inference (NLI)~\cite{rte1,rte2,rte3,rte5,winograd2012,mnli2018}. The diversity of the tasks makes GLUE very suitable for evaluating the generalization and robustness of NLU models. 

\noindent $\bullet$ \textbf{SNLI}.
The Stanford Natural Language Inference (SNLI) dataset contains 570k human annotated sentence pairs, in which the premises are drawn from the captions of the Flickr30 corpus and hypotheses are manually annotated \cite{snli2015}. 
This is the most widely used entailment dataset for NLI.

\noindent $\bullet$ \textbf{SciTail}.
This is a textual entailment dataset derived from a science question answering (SciQ) dataset \cite{scitail}. The task involves assessing whether a given premise entails a given hypothesis.  
In contrast to other entailment datasets mentioned previously, the hypotheses in SciTail are created from science questions while the corresponding answer candidates and premises come from relevant web sentences retrieved from a large corpus. As a result, these sentences are linguistically challenging and the lexical similarity of premise and hypothesis is often high, thus making SciTail particularly difficult. 

\noindent $\bullet$ \textbf{MedNLI}. 
This is a textual entailment dataset in the clinical domain. It was derived from medical history of patients and annotated by doctors. The task involves assessing whether a given premise entails a given hypothesis. The hypothesis sentences in this dataset were generated by clinicians, while corresponding answer candidates and premises come from MIMIC-III v1.3 \cite{johnson2016mimic}, a database containing 2,078,705 clinical notes written by healthcare professionals. Its specialized domain nature makes MedNLI a challenging dataset.

\noindent $\bullet$ \textbf{HANS}.
This is an NLI evaluation set that tests three hypotheses about invalid heuristics that NLI models are likely to learn: lexical overlap (assume that a premise entails all hypotheses constructed from words in the premise), subsequence (assume that a premise entails all of its contiguous subsequences), and constituent. HANS is a challenging dataset that aims to test how much models are vulnerable to such heuristics, and standard training often results in models failing catastrophically, even models such as BERT \cite{mccoy2019hans}. 
%HANS is a template-generated challenge set designed to test whether NLI models have adopted three syntactic heuristics: 

\noindent $\bullet$ \textbf{XNLI}.
This is a cross-lingual natural language inference dataset built by extending the development and test sets of the Multi-Genre Natural Language Inference Corpus \cite{mnli2018} to 15 languages, including low-resource languages such as Swahili. This corpus was designed to evaluate cross-language sentence understanding, where models are supposed to be trained in one language and tested in different ones. Validation and test sets are translated from English to 14 languages by professional translators, making results across different languages directly comparable \cite{artetxe2019massively}.

% \section{Implementation Details}
% \label{sec:param}
% Our implementation is based on the MT-DNN toolkit \cite{mtdnn2020demo}. We follow the default hyperparameters used for fine-tuning the uncased BERT base model \cite{devlin2018bert, mtdnn2020demo}. 
% Specifically, we use $0.1$ for the dropout rate except 0.2 for MNLI, $0.01$ for the weight decay rate and the Adamax \cite{kingma2014adam} optimizer with the default Lookahead \cite{zhang2019lookahead} to stabilize training. 
% We select the learning rate from $\{5\mathrm{e}{-5}, 1\mathrm{e}{-4}\}$ for all the models. The maximum training epoch is set to 6, and the we follow \cite{jiang2019smart} to set adversarial training hyperparameters: $\epsilon=1e-5$ and $\eta=1e-4$. In our experiments, we simply set $\alpha=1$ in Eq 1.
 
% \section{Loss Landscape}
%  \label{sec:lossland}
%  We also visualize the loss landscape of both the standard training and {\DNAME}, shown in Figure~\ref{fig:loss-surface}. {\DNAME} has a wider and flatter loss surface indicating a better generalization \cite{hochreiter1997flat,hao2019vbert,li2018landscape}. 

%% file: tables/dataset.tex
\begin{table*}[!htb]
	\begin{center}
		\begin{tabular}{l|l|c|c|c|c|c}
			\toprule 
			\bf Corpus &Task& \#Train & \#Dev & \#Test   & \#Label &Metrics\\ \hline \hline
			\multicolumn{6}{@{\hskip1pt}r@{\hskip1pt}}{Single-Sentence Classification (GLUE)} \\ \hline
			CoLA & Acceptability&8.5k & 1k & 1k & 2 & Matthews corr\\ \hline
			SST & Sentiment&67k & 872 & 1.8k & 2 & Accuracy\\ \hline \hline
			\multicolumn{6}{@{\hskip1pt}r@{\hskip1pt}}{Pairwise Text Classification (GLUE)} \\ \hline
			MNLI & NLI& 393k& 20k & 20k& 3 & Accuracy\\ \hline
            RTE & NLI &2.5k & 276 & 3k & 2 & Accuracy \\ \hline
            WNLI & NLI &634& 71& 146& 2 & Accuracy \\ \hline
			QQP & Paraphrase&364k & 40k & 391k& 2 & Accuracy/F1\\ \hline
            MRPC & Paraphrase &3.7k & 408 & 1.7k& 2&Accuracy/F1\\ \hline
			QNLI & QA/NLI& 108k &5.7k&5.7k&2& Accuracy\\ \hline \hline
			\multicolumn{5}{@{\hskip1pt}r@{\hskip1pt}}{Text Similarity (GLUE)} \\ \hline
			STS-B & Similarity &7k &1.5k& 1.4k &1 & Pearson/Spearman corr\\ \hline
			\multicolumn{6}{@{\hskip1pt}r@{\hskip1pt}}{Pairwise Text Classification for the Zero-shot setting} \\ \hline
			SNLI & NLI&  - &-&9.8k&3& Accuracy\\ \hline
			SciTail & NLI& - &-&2.1k&2& Accuracy\\ \hline
			HANS & NLI& - &-&3k&2& Accuracy\\ \hline
			MedNLI & NLI& - & - &1.4k&3& Accuracy\\ \hline
			XNLI & NLI&- &-&75k&3& Accuracy\\ 
			\bottomrule

		\end{tabular}
	\end{center}
	\caption{Summary information of the NLU benchmarks. 
	}
	\label{tab:datasets}
\end{table*}